\documentclass[environsciproc,article,submit,pdftex,moreauthors]{Definitions/mdpi} 

\preto{\abstractkeywords}{\nolinenumbers}

\usepackage{pgffor}
\usepackage{subfig}
\usepackage{comment}

\firstpage{1} 
\pubvolume{1}
\issuenum{1}
\articlenumber{0}
\pubyear{2023}
\copyrightyear{2023}
\datereceived{ } 
\daterevised{ } 
\dateaccepted{ } 
\datepublished{ } 
\hreflink{https://doi.org/} 

\Title{Zero-Shot Refinement of Buildings' Segmentation Models using SAM}

\TitleCitation{Zero-Shot Refinement of Buildings' Segmentation Models using SAM }

\Author{Ali Mayladan$^{1,2}$, Hasan Nasrallah$^{2}$, Hasan Moughnieh$^{2}$, Mustafa Shukor$^{3}$ and Ali J. Ghandour\orcidA{} $^{2}$*}

\AuthorNames{Ali Mayladan, Hasan Nasrallah, Hasan Moughnieh, Mustafa Shukor and Ali J. Ghandour}

\AuthorCitation{Mayladan, A.; Nasrallah, H.; Moughnieh, H.; Shukor M.; Ghandour A. J.}

\address{%
$^{1}$ \quad Lebanese University;\\
$^{2}$ \quad National Center for Remote Sensing - CNRS, Lebanon;\\
$^{3}$ \quad Sorbonne University;}

\corres{Corresponding author: Ali J. Ghandour, aghandour@cnrs.edu.lb;}

\abstract{
Foundation models have excelled in various tasks but are often evaluated on general benchmarks. The adaptation of these models for specific domains, such as remote sensing imagery, remains an underexplored area. In remote sensing, precise building instance segmentation is vital for applications like urban planning. While Convolutional Neural Networks (CNNs) perform well, their generalization can be limited.
For this aim, we present a novel approach to adapt foundation models to address existing models' generalization dropback. Among several models, our focus centers on the Segment Anything Model (SAM), a potent foundation model renowned for its prowess in class-agnostic image segmentation capabilities. We start by identifying the limitations of SAM, revealing its suboptimal performance when applied to remote sensing imagery. Moreover, SAM does not offer recognition abilities and thus fails to classify and tag localized objects. To address these limitations, we introduce different prompting strategies, including integrating a pre-trained CNN as a prompt generator. This novel approach augments SAM with recognition abilities, a first of its kind. We evaluated our method on three remote sensing datasets, including the WHU Buildings dataset, the Massachusetts Buildings dataset, and the AICrowd Mapping Challenge. For out-of-distribution performance on the WHU dataset, we achieve a 5.47\% increase in IoU and a 4.81\% improvement in F1-score. For in-distribution performance on the WHU dataset, we observe a 2.72\%  and 1.58\% increase in True-Positive-IoU and True-Positive-F1 score, respectively. Our code is publicly available at this \href{https://github.com/geoaigroup/GEOAI-ECRS2023}{Repo}, hoping to inspire further exploration of foundation models for domain-specific tasks within the remote sensing community.
}

\begin{document}

\section{Introduction}

\begin{figure}
    \centering
    \includegraphics[scale=0.3]{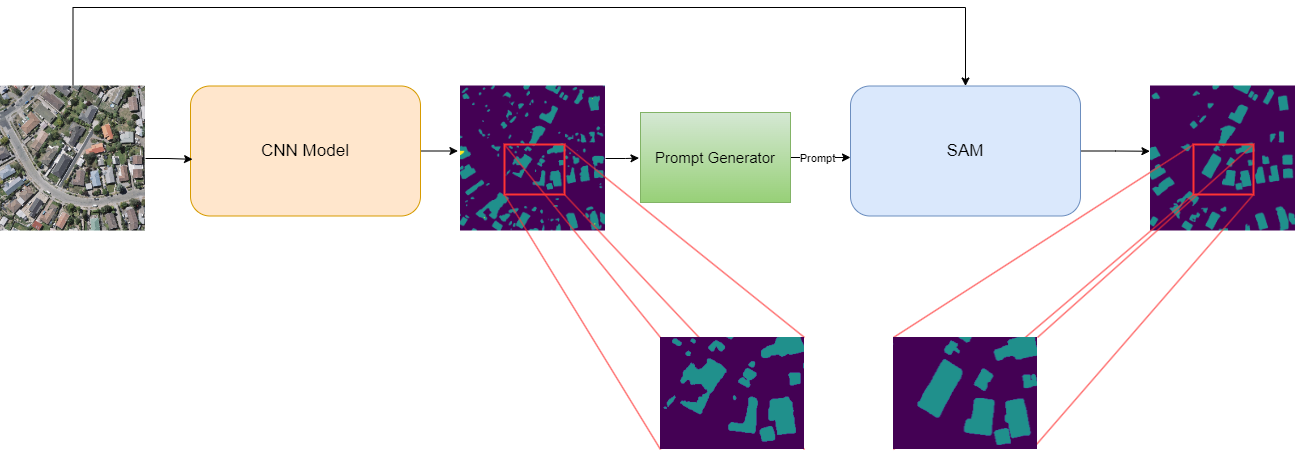}
    \caption{Input RGB image undergoes rooftop instance segmentation via the CNN model. Segmentation masks are passed to the Prompt Generator used to prompt SAM. This approach would equip SAM with recognition abilities and generate precise buildings output masks.}
    \label{fig:model}
\end{figure}

Most current state-of-the-art remote sensing models are CNN-based~\cite{oshea2015introduction} that struggle with out-of-distribution generalization. This challenge is mainly due to the significant variations in imagery when observed in various regions, seasons, and periods. This demonstrates the need for more robust and adaptive techniques to manage these variances efficiently. 

Foundation models~\cite{awais2023foundational} have demonstrated unparalleled proficiency in a wide range of tasks, from high-resolution image interpretation to multi-modal data analysis. These models not only have equaled, but frequently outperformed, the performance of previous task-focused algorithms, particularly in complex tasks such as dense prediction and spatial pattern recognition~\cite{ZUO2022109552}. However, many of these models have been trained and benchmarked primarily against generic datasets and usually underperform on domain-specific tasks like remote sensing segmentation. 

The use of these foundation models for remote sensing applications such as land cover categorization, change detection, and instance segmentation is still unexplored. Therefore, the adaptation of foundational models is crucial to address the evolving challenges in satellite and aerial data analysis.

A central question then arises: \textbf{How can foundation models be effectively leveraged for remote sensing segmentation, notably, in buildings' footprint instance segmentation?}

We mainly focus in this manuscript on Meta's newly unveiled transformer-based "Segment Anything Model" (SAM)~\cite{kirillov2023segment}, a powerful foundation model for image segmentation, promising broad applicability and high accuracy. SAM is trained on an extensive high-quality dataset (SA-1B) encompassing more than 11 million images and more than 1.1 billion masks, constituting the largest segmentation dataset, with 400x more masks than any existing segmentation dataset. 

Although SAM excels in localization capabilities, it does not offer recognition abilities and thus fails to classify and tag localized objects. Therefore, the use of SAM for segmentation is not a straightforward task. Hence, we propose to leverage the SAM foundation model to improve the performance of pre-trained CNN segmentation models. Complementing CNNs with SAM might harness: (\textit{i}) the collaboration between CNNs and transformers, on the one hand, and (\textit{ii}) the generalization power of SAM with the domain specificity of pre-trained CNNs. Specifically, we propose to Prompt Engineer (PE) SAM to enhance its performance by integrating a pre-trained CNN as a prompt generator.

Our contributions can be summarized as follows: (\textit{i}) Investigate various distinct single and composite prompting strategies for SAM and (\textit{ii}) Experiment with two CNN models as prompt generators for SAM to get more accurate instance segmentation results.

\section{Related Work}
In remote sensing, instance segmentation is vital for precisely extracting building footprints from satellite and aerial imagery, allowing for exact land-use analysis and infrastructure planning. It provides real-time insights for various geographic tasks by differentiating between individual items of the same class.

Foundation models have emerged as a transformational force in the ever-evolving field of artificial intelligence. Among these, large language models (LLMs) ~\cite{openai2023gpt4,touvron2023llama} have demonstrated unparalleled capabilities in natural language processing and generation, enabling a wide range of applications ranging from chatbots to content development.

On another front, multimodality-based models~\cite{shukor2023unifiedunival,liu2023grounding} can analyze and integrate data from various modalities such as images, text, and sound. Within the visual domain, the foundation models ~\cite{he2015deep,wang2023seggpt,liu2021swin} have set new standards for image recognition, object detection, and various other computer vision tasks. These models have served as the backbone for many applications, ranging from autonomous vehicles to healthcare diagnostics.

Leveraging and adapting foundation models to specialized tasks has been a notable trend in recent research~\cite{shukor2023epalm,zhang2023textguided}. 
SAM, specifically, has been widely deployed in a short period of time for various applications ~\cite{wu2023medical,osco2023segment,ding2023adapting, zhang2023personalize}. With a strong Zero-Shot performance, SAM has attracted attention for its outstanding capacity to generate high-quality object masks from a variety of input prompts.

\section{SAM Prompt Engineering}

\begin{table}[h!]
    \centering
    \resizebox{\textwidth}{!}{%
        \begin{tabular}{ |c|c c c c || c c| } 
         \hline
         Experiment & Precision & Recall& IoU & F1 &  TP-IoU & TP-F1 \\
         \hline\hline
         Single-point & 69.40 & 63.63 & 47.21 & 51.83&  81.12 & 89.05  \\ 
         Single-point + Negative-point & 72.31 & 66.52 & 50.62 & 55.36 & 81.89 & 89.56 \\
         \hline
         Skeleton Multiple-points & 83.18 & 76.97 & 60.97 & 65.96 & 84.15 & 91.03 \\
         Random Multiple-points & 84.12 & 78.01& 61.52 & 66.64 &  83.89 & 90.88 \\
         Random Multiple-points + Single-point & 84.09 & 78.04&  61.86 & 67.00 &  83.92 & 90.89 \\
         Random Multiple-points + Negative-point & 83.72 & 77.68& 61.12 & 66.23 &  83.79  & 90.81 \\
         \hline
         Bounding-box &  84.78 & 78.62 & 63.82 & 68.52 & \textbf{85.67}  & \textbf{91.98} \\
         Bounding-box + Single-point & \textbf{84.88} & 78.72 & \textbf{63.86} & \textbf{68.62} &   85.53 & 91.90  \\
         Bounding-box + Multiple-points & 84.87 & \textbf{78.81} & 63.54 & 68.43 &  85.10 & 91.65  \\
        [1ex] 
         \hline
          baseline U-Net-based CNN~\cite{nasrallah2022lebanon}& 84.76 & 78.68 & 61.79 & 67.34 & 82.95 & 90.40 \\
          \hline
        \end{tabular}
    }
    \captionsetup{font=footnotesize}
    \caption{Comprehensive set of experiments conducted while integrating the U-Net CNN model~\cite{nasrallah2022lebanon} with SAM on WHU Buildings' dataset, encompassing various prompt types. These experiments are evaluated based on precision, recall, IoU, F1-score, True-Positive IoU (TP-IoU) and True-Positive F1-score (TP-F1) metrics.}
    \label{tab:Unet}
\end{table}

Meta AI recently unveiled the Segment Anything Model (SAM)~\cite{kirillov2023segment}, a class-agnostic segmentation model that incorporates automatic mask generation and quality filters. SAM utilizes a Vision Transformer (ViT) for image encoding and employs a two-layer mask decoder with transformer-based architecture. SAM has outstnading localization capabilities but lacks any recognition abilities.

In our proposed methodology shown in Figure \ref{fig:model}, we augmnented SAM with the capability to recognize objects, mainly buildings. The input image is fed initially to a CNN-based model pre-trained for buildings' instance segmentation. We then developed a prompt generator component capable of providing SAM with various promptst. At the core of our proposed architecture lies this prompt generator, which operates by taking the output masks of the CNN model as input and using them to generate SAM prompts of the following three different categories: (\textit{i}) single-point prompts, where a single representative point is generated for each input mask. (\textit{ii}) Multiple-point prompts, using either random points localized within the input mask or by extracting skeleton-shaped points from the input buildings' mask. (\textit{iii}) Bounding box prompts for each mask, with the box coordinates serving as prompts for SAM. 

The proposed component can also generate hybrid prompts of various categories such as a "single-point and bounding-box" prompt. We also used the concept of negative points that can be located either in the image background or inside the bounding box, but not within the building's mask. More details about the three prompt categories are presented in Table~\ref{tab:Unet}.

We experimented with two different CNN models trained for buildings' footprints instance segmentation: (\textit{i}) a U-Net-based CNN model~\cite{nasrallah2022lebanon} that employs Efficient-Net-B3 backbone for feature extraction, ensuring accuracy and precision in mask generation, and (\textit{i}) D-LinkNet~\cite{Zhou_2018_CVPR_Workshops} that builds upon LinkNet and utilizes ResNet34 as encoder. The encoder includes dilated convolution layers for context capture, and the decoder efficiently restores feature map resolution through transposed convolution layers. More details of these experiments are elaborated on in the next section.

\section{Experimental Results}
In this section, we provide insights into our dual-architecture shown in Figure \ref{fig:model} that consists of a CNN prompt generator and SAM for Zero-Shot mask refinement. 

We conducted a series of experiments to assess the performance of SAM under various types of prompts in three remote sensing datasets: the WHU Buildings dataset~\cite{8444434} and the Massachusetts Buildings dataset~\cite{MnihThesis}, in addition to the AICrowd Mapping Challenge dataset~\cite{mohanty2020deep}. We consistently used prediction masks generated by one of the two CNN models as input to the prompt generator.

\begin{figure*}[t]
    \centering
    \subfloat[Single-point]{\includegraphics[scale=0.2]{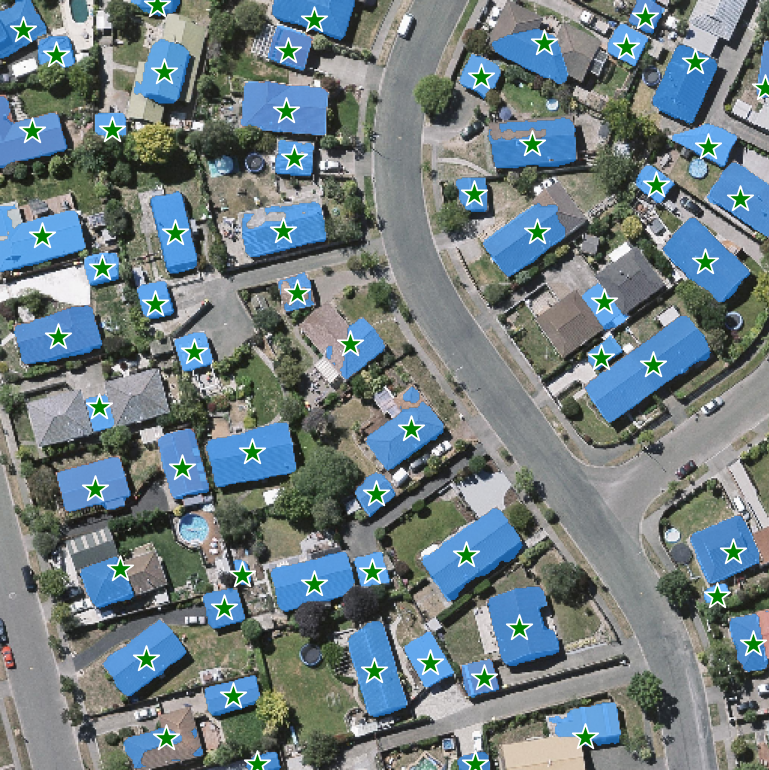}}\hspace{1pt}
    {\includegraphics[scale=0.2]{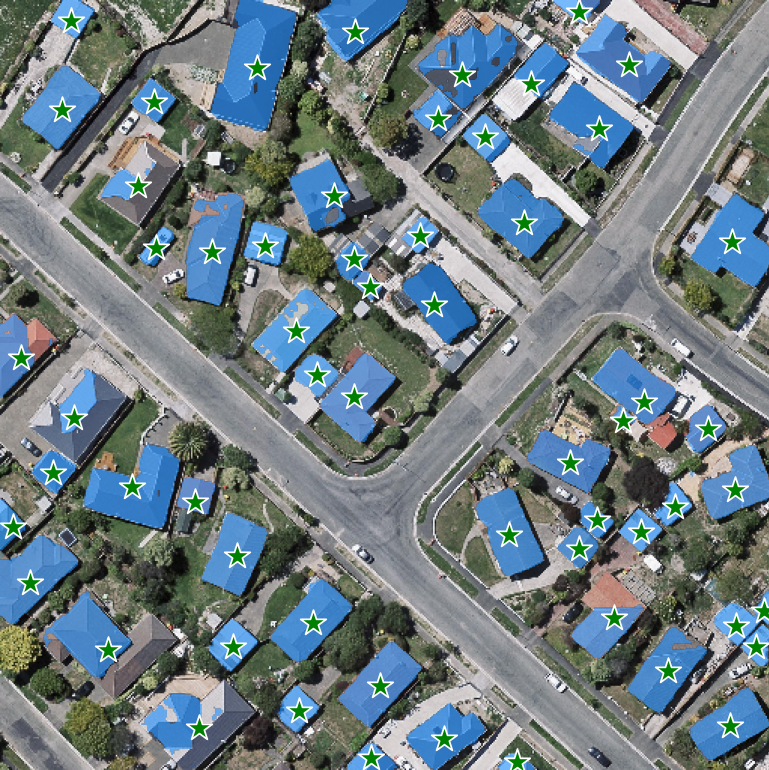}}\hspace{1pt}
    {\includegraphics[scale=0.2]{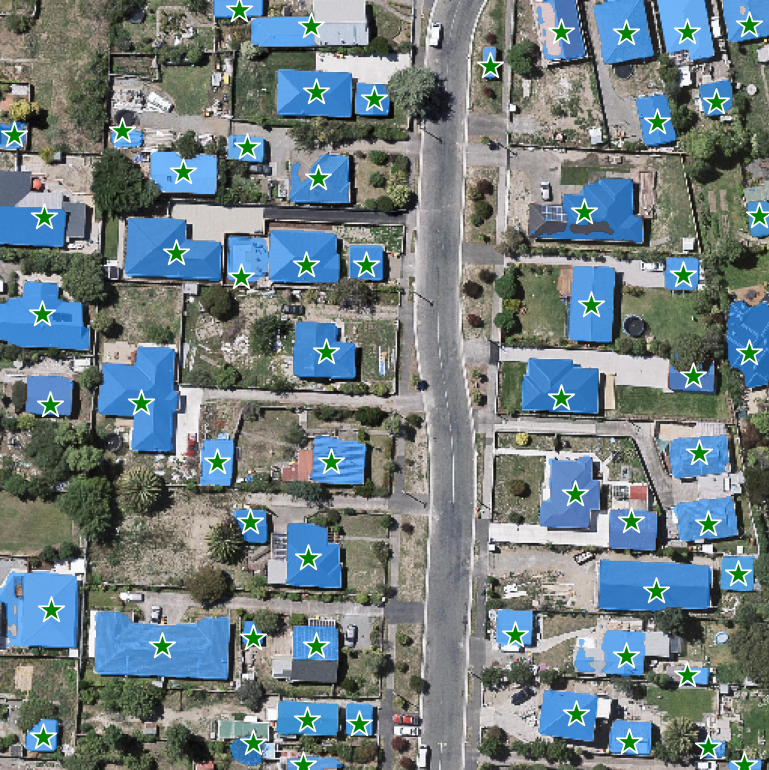}} \\ 

    \subfloat[Single + Negative-points]{\includegraphics[scale=0.2]{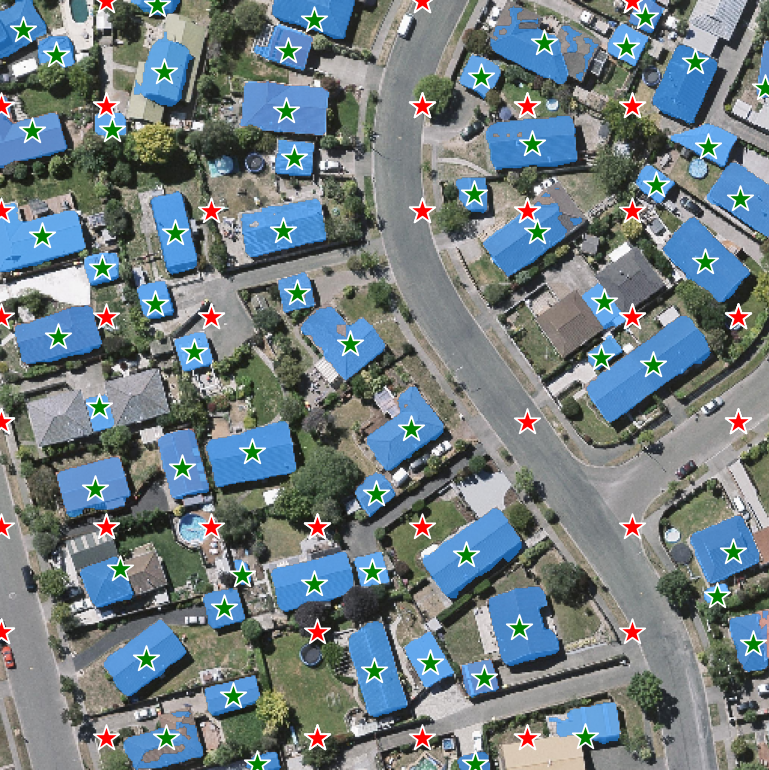}}\hspace{1pt}
    {\includegraphics[scale=0.2]{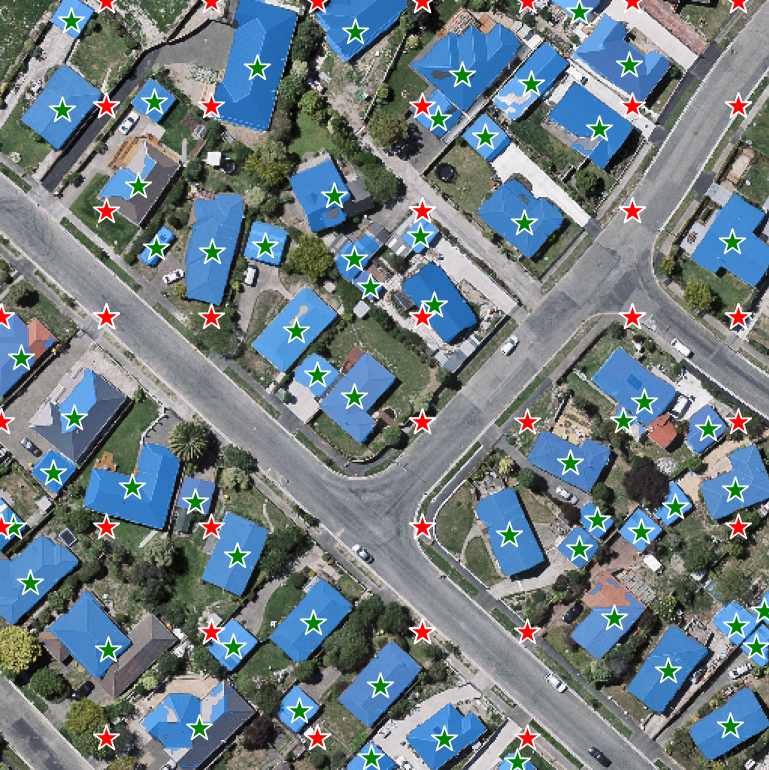}}\hspace{1pt}
    {\includegraphics[scale=0.2]{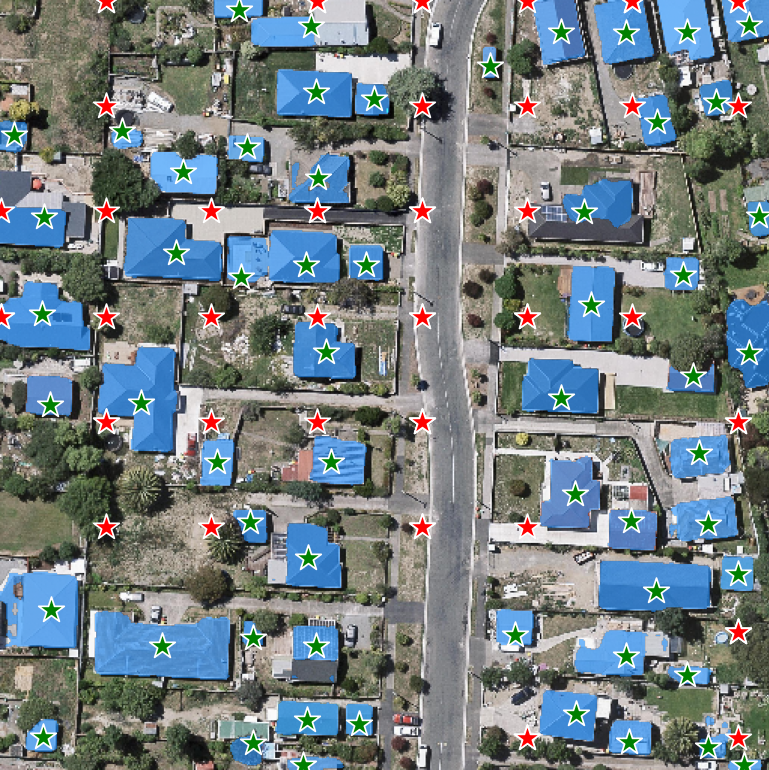}} \\

    \subfloat[Skeleton Multiple-points]{\includegraphics[scale=0.2]{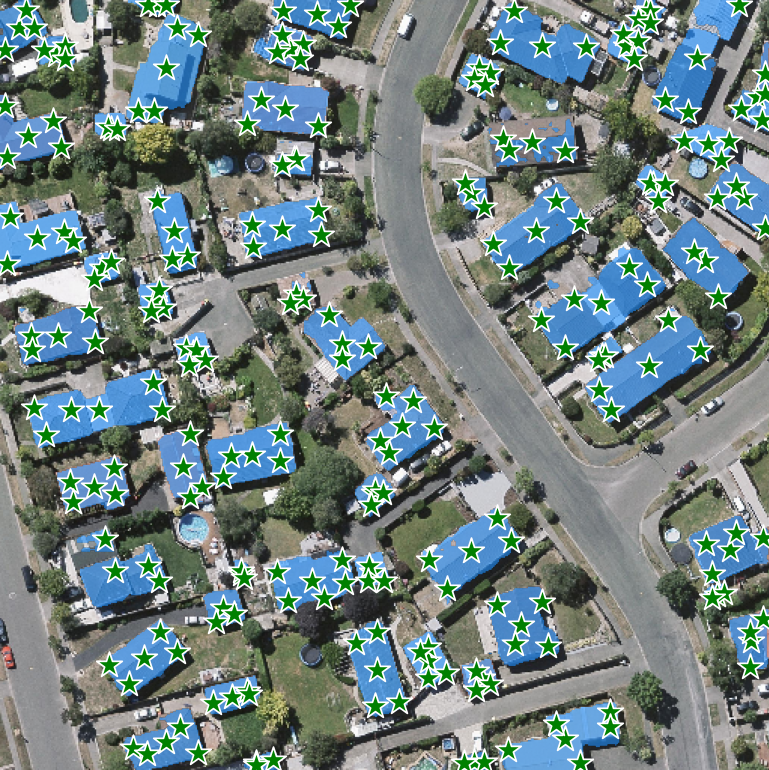}}\hspace{1pt}
    {\includegraphics[scale=0.2]{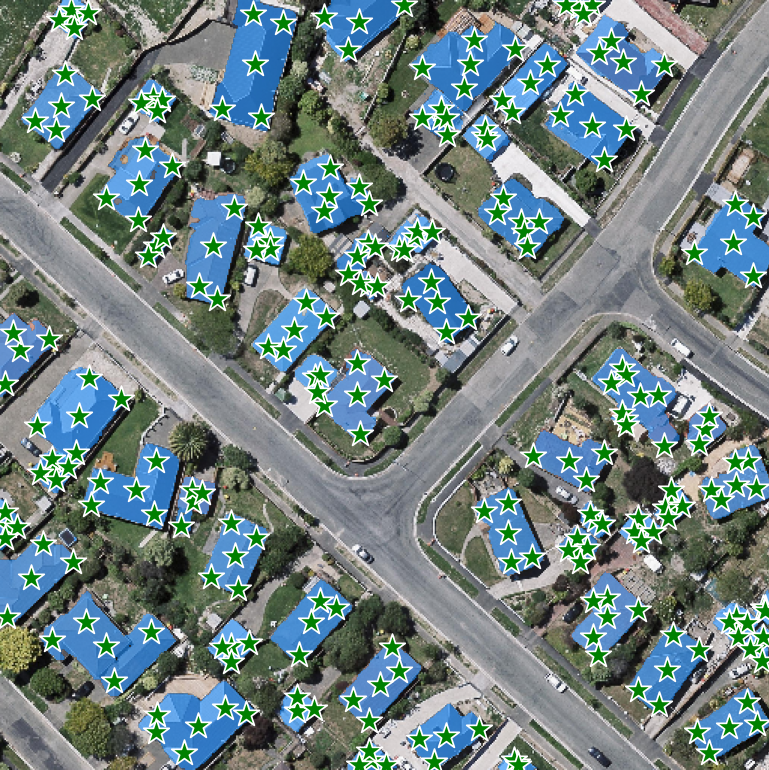}}\hspace{1pt}
    {\includegraphics[scale=0.2]{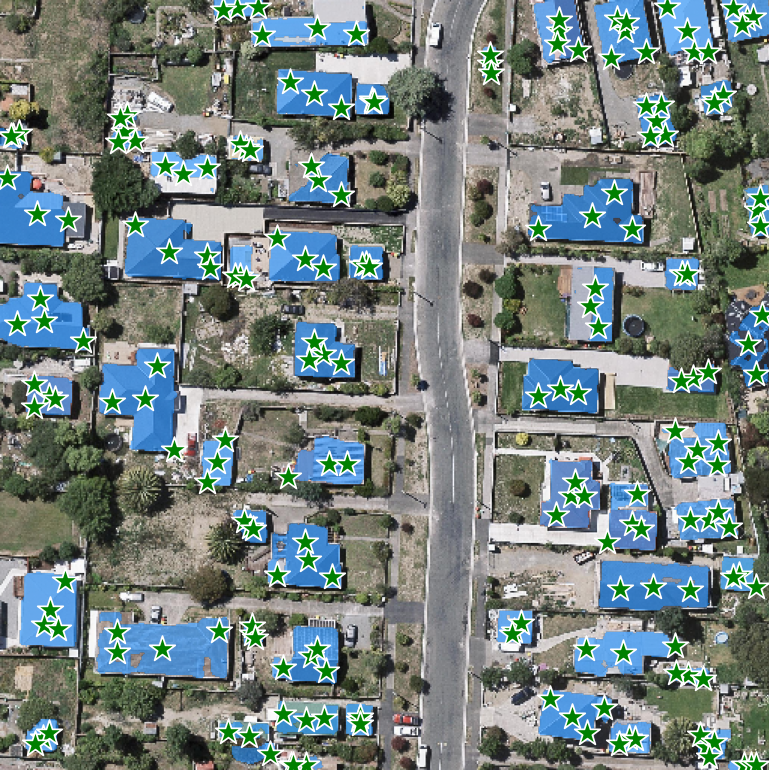}} \\

    \subfloat[Random Multiple-points]{\includegraphics[scale=0.2]{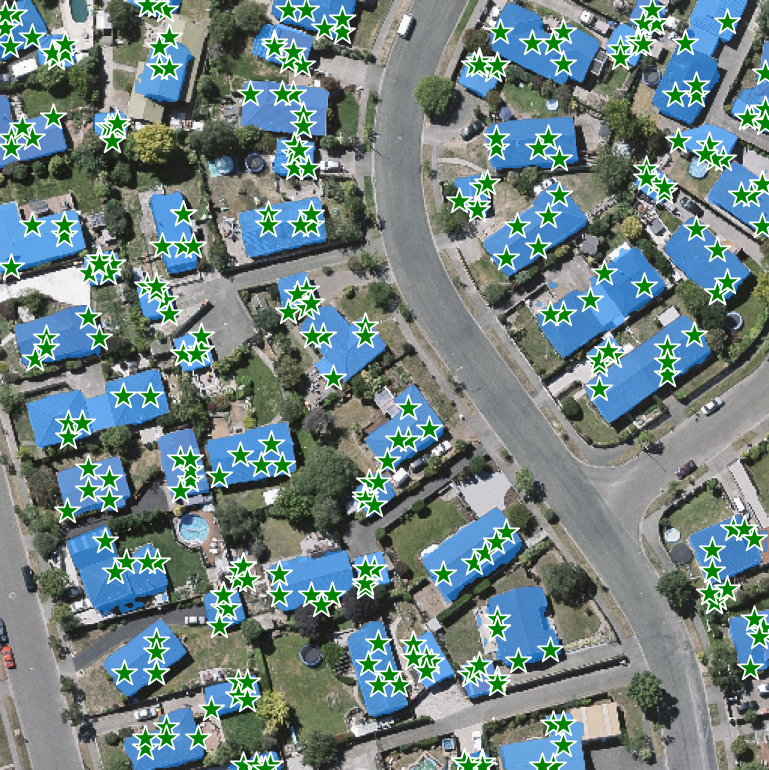}}\hspace{1pt}
    {\includegraphics[scale=0.2]{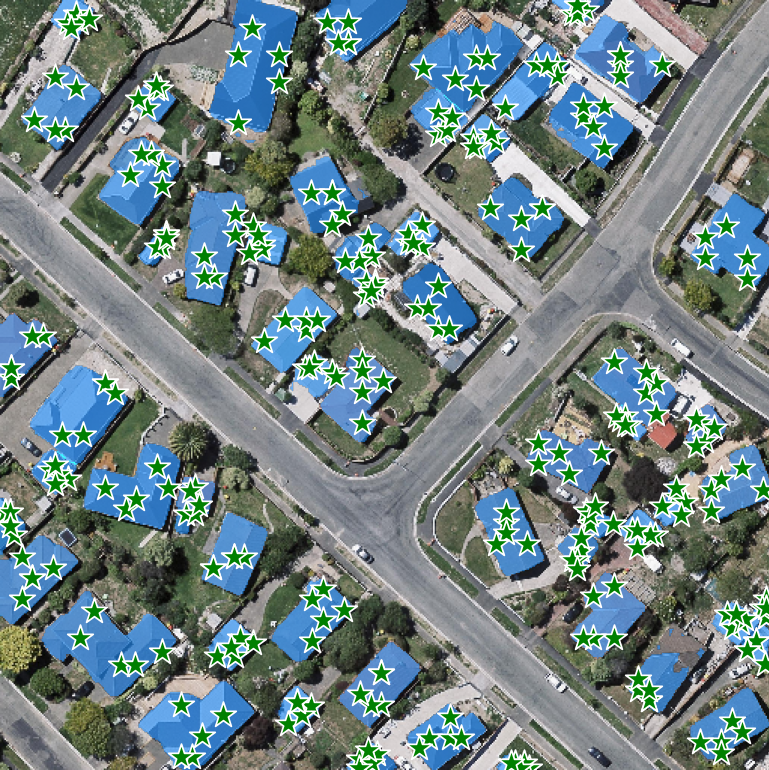}}\hspace{1pt}
    {\includegraphics[scale=0.2]{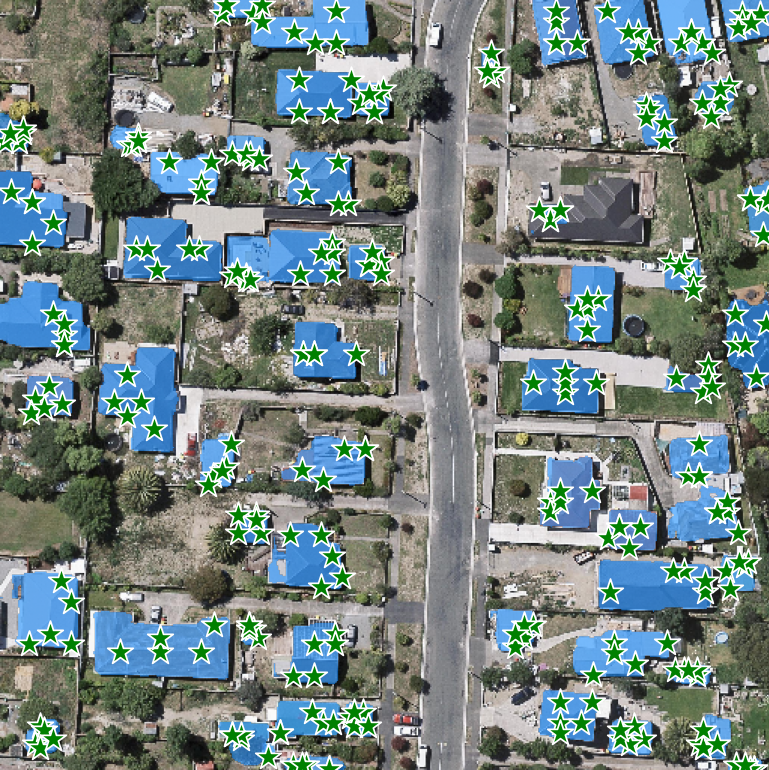}} \\

    \subfloat[Bounding-box]{\includegraphics[scale=0.2]{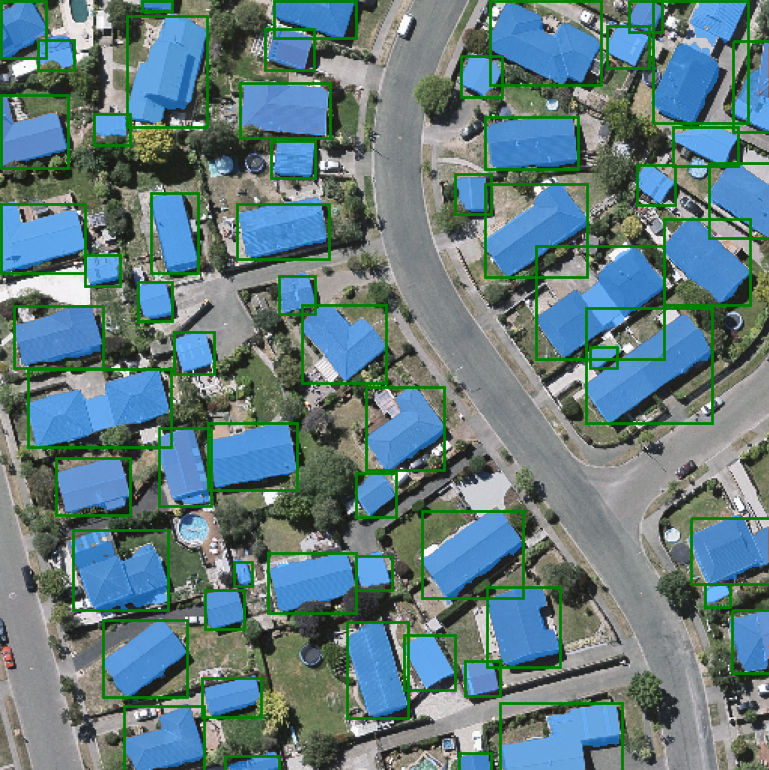}}\hspace{1pt}
    {\includegraphics[scale=0.2]{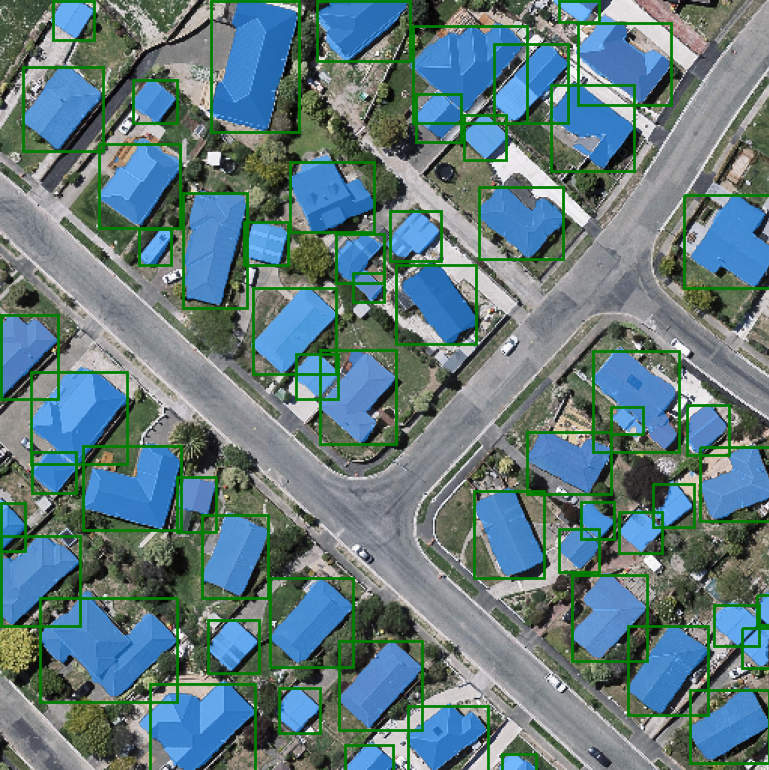}}\hspace{1pt}
    {\includegraphics[scale=0.2]{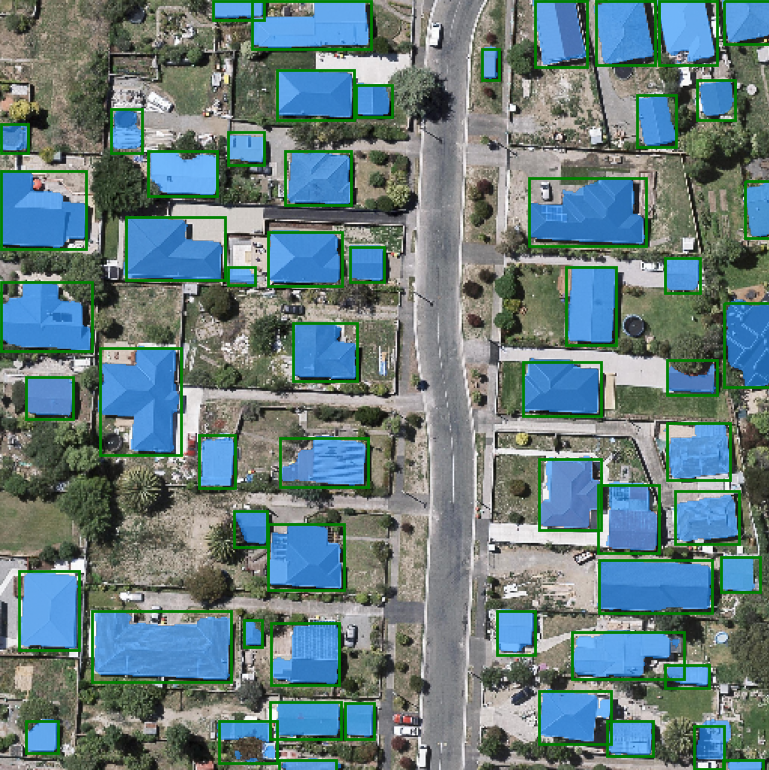}} 
    
    \caption{Visualizations, over three different images from WHU dataset, of prompt engineering experiments including Single-point, Single-point + Negative-points (in red), Skeleton Multiple-points, Random Multiple-points and Bounding-box.}
\label{fig:whu_pe}
\end{figure*}

As detailed in Table \ref{tab:Unet}, we initially used single-point prompts for buildings' rooftop instance segmentation. We replace each CNN-predicted building's mask by a single-representative point, which is then provided as input to SAM alongside the original RGB image. The representative point, by definition, is guaranteed to be within the building, irrespective of the building's shape. The SAM output in Figure~\ref{fig:whu_pe}(a) reveals that the representative single-point sometimes fails to accurately segment the target object due to irregular shapes (e.g., L-shaped or U-shaped buildings). Single-point prompt scores 47.21\%, 51.83\%, 81.12\% and 89.05\%, in terms of IoU, F1, TP-IoU, and TP-F1 scores, respectively, on the WHU dataset. Using one Negative-point along the single-point improved IoU and F1-score with more than 3\%. 

We explored the use of multiple-points prompts, to ensure comprehensive coverage of the building area, and improve segmentation accuracy, particularly for larger buildings where a single-point prompt might be insufficient to encompass the entire structure. We distribute 5 points within each mask following two approaches: (\textit{i}) Random distribution as shown in Figure~\ref{fig:whu_pe}(d) and (\textit{ii}) Skeleton form depicted in Figure~~\ref{fig:whu_pe}(c) where one point is the building centroid and the others along the edges. Surprisingly, Table~\ref{tab:Unet} reveals that both Random and SkeletonMultiple-points prompt almost exhibit the same performance. Future research is needed to investigate why the Skeleton approach did not outperform the random scheme. We also conducted additional pairs of experiments using "RandomMultiple-points + Single-point" and "RandomMultiple-points + Negative-point" where both did not provide substantial improvement.

Among all the experiments depicted in Table~\ref{tab:Unet}, the Bounding-box prompts exhibited the most promising results. We explored three difference version including: (\textit{i}) Bounding-box, (\textit{ii}) Bounding-box + Single-point and (\textit{ii}) Bounding-box +Multiple-points. The prompt of SAM with Bounding-box led to 2.03\%, 1.18\%, 2.72\% and  1.58\% improvement in terms of IoU, F1, TP-IoU, and TP-F1 scores, respectively, on the WHU dataset.

We also performed prompt engineering experiments with bounding-boxes using D-LinkNet CNN~\cite{Zhou_2018_CVPR_Workshops} out-of-distribution on the WHU dataset. Using bounding-box prompt showed substantial improvement with a \textbf{5.47\%}, \textbf{4.81\%}, \textbf{7.56\%} and \textbf{4.71\%} increase in IoU, F1-score, TP-IoU and TP-F1-score, respectively, on the WHU dataset as shown in Table~\ref{tab:Dlinknetwhu}.

Additionally, we expanded our experiments to include the Massachusetts buildings and the AICrowd Mapping Challenges datasets using bounding boxes as prompts. On the AICrowd dataset, SAM proficiently predicts building segments, even those obscured by trees, while ground truth designates tree-covered sections as integral building parts. Similarly, on the Massachusetts Buildings dataset, we noticed improvements in terms of TP-IoU and TP-F1. Detailed results over these two datasets are omitted for space limitations.

\begin{table}[]
    \centering
    \resizebox{\textwidth}{!}{%
    \begin{tabular}{ |c|c c c c || c c| } 
     \hline
     Experiment & Precision & Recall& IoU & F1 &  TP-IoU & TP-F1 \\
     \hline\hline
     Bounding-box & \textbf{43.25} & \textbf{52.16} & \textbf{29.96} & \textbf{32.44} & \textbf{83.72} & \textbf{90.64} \\
     \hline
     baseline DCNN~\cite{Zhou_2018_CVPR_Workshops} & 39.89 & 47.59 & 24.49 & 27.63 & 76.16 & 85.93\\
     \hline
    \end{tabular}
    }
    \captionsetup{font=footnotesize}
    \caption{SAM Bouding-box prompt results using the D-LinkNet mode~\cite{Zhou_2018_CVPR_Workshops} out-of-distribution on the WHU Buildings dataset.
    }
    \label{tab:Dlinknetwhu}
\end{table}

\section{Conclusion}

In this paper, we propose to leverage SAM model in the domain of building segmentation for remote sensing applications. Our approach introduces a novel adaptation paradigm based on prompting, where we exploit the power of a pre-trained CNN as a prompt generator. We conduct an extensive evaluation of our approach on the WHU dataset, yielding remarkable improvements in SAM's building segmentation accuracy. In the context of out-of-distribution performance, our results demonstrated an impressive boost, with a notable 5.47\% enhancement in IoU and a substantial 4.81\% improvement in F1-score. Moreover, our evaluation also revealed noteworthy enhancements for in-distribution performance on the WHU dataset, showcasing a 2.72\% increase in True-Positive-IoU and a significant 1.58\% enhancement in True-Positive-F1-score. These results underline the effectiveness of our method in diverse scenarios. We hope this work will inspire the broader academic community to explore the potential of foundation models for domain-specific tasks. 

\vspace{6pt} 




 \funding{``This research received no external funding''.}

 \conflictsofinterest{``The authors declare no conflict of interest.''}

\begin{adjustwidth}{-\extralength}{0cm}

\reftitle{References}


\bibliography{references_updated}


\end{adjustwidth}
\end{document}